%% file: main.tex
\documentclass[10pt,twocolumn,letterpaper]{article}

\usepackage[pagenumbers]{cvpr} % To produce the CAMERA-READY version
\input{preamble}

\definecolor{cvprblue}{rgb}{0.21,0.49,0.74}
\usepackage[pagebackref,breaklinks,colorlinks,allcolors=cvprblue]{hyperref}

\def\paperID{*****} % *** Enter the Paper ID here
\def\confName{CVPR}
\def\confYear{2026}

\title{Seeing Is Not Measuring: Tool-Augmented Metric Spatial Reasoning for Vision-Language Models}

\author{Clemens Grange \quad Kai Glantz\\
Technical University of Munich\\
{\tt\small \{clemens.grange, kai.glantz\}@tum.de}
}

\begin{document}
\maketitle
% ---------------------------------------------------------------------------
% PAGE BUDGET (4 pages max for main text; references do NOT count).
%   Abstract + Intro ........ ~0.75 col-page
%   Related Work ............ ~0.5
%   Method (+ pipeline fig) . ~1.0
%   Experiments (tables/figs) ~1.25
%   Conclusion + Next Steps . ~0.4
% Appendix (X_suppl) = optional extra visuals only; report must stand alone.
% ---------------------------------------------------------------------------
\input{sec/0_abstract}
\input{sec/1_intro}
\input{sec/2_related}
\input{sec/3_method}
\input{sec/4_experiments}
\input{sec/5_conclusion}
\input{sec/6_acknowledgments}
{
    \small
    \bibliographystyle{ieeenat_fullname}
    \bibliography{main}
}

% WARNING: do not forget to delete the supplementary pages from your submission
\input{sec/X_suppl}

\end{document}

%% file: preamble.tex
\usepackage{microtype}
\usepackage{placeins}
\usepackage{float}   % provides [H]: pin a float exactly where it is declared

\usepackage{listings}
\lstdefinestyle{prompt}{%
  basicstyle=\ttfamily\scriptsize,
  breaklines=true,
  breakatwhitespace=true,
  breakindent=0pt,
  breakautoindent=false,
  postbreak=\mbox{\textcolor{gray}{$\hookrightarrow$}\space},
  columns=fullflexible,
  keepspaces=true,
  showstringspaces=false,
  frame=single,
  framesep=4pt,
  xleftmargin=5pt,
  xrightmargin=3pt,
  aboveskip=5pt,
  belowskip=3pt,
  numbers=none,
}

%% file: sec/0_abstract.tex
\begin{abstract}
Vision-Language Models (VLMs) describe scenes well but reason poorly about metric 3D structure such as absolute distances, physical sizes, or egocentric directions. We present a modular, predictor-agnostic, tool-augmented framework that equips a small VLM (Qwen3.5-4B) with geometric tools: 3D object detection, metric depth estimation, and deterministic solvers for distance, size and bearing. Each object is detected in the camera frame of its own best view, and the tools use that frame's pose to lift every detection into one shared world frame. Moving metric computation out of the model's weights and into explicit solvers yields large gains on three of four ReVSI-Bench tasks: with a strong monocular detector (WildDet3D), absolute distance rises from $0.46$ to $0.74$ Mean Relative Accuracy (MRA), relative distance from $39.1\%$ to $67.4\%$, and relative direction from a below-chance $25.9\%$ to $73.4\%$. Because any detector can be swapped in behind the tool interface, comparing real detectors against ground-truth boxes separates perception error from reasoning error: orchestration costs only $0.03$ MRA. Object size is bounded by the detector: the tools are near-exact on ground-truth boxes ($0.97$) yet the best real detector barely beats the no-tool baseline ($0.61$ vs.\ $0.58$), because size reads straight off a box extent monocular detectors get wrong. Without a predefined recipe, the model already sequences the tools correctly on its own, matching a scripted pipeline on three of four tasks.
\end{abstract}

%% file: sec/1_intro.tex
\section{Introduction}
\label{sec:intro}

Embodied and robotic agents operating in real environments need precise metric spatial understanding: how far apart two objects are, how large they are, and where they lie relative to the viewer. Vision-Language Models (VLMs) excel at qualitative scene description, yet remain unreliable at exactly this kind of metric 3D reasoning.

% Declared after the first paragraph so the float lands at the top of the RIGHT
% column (next available column top) rather than above the abstract on the left.
\begin{figure}[t]
  \centering
  \includegraphics[width=\linewidth]{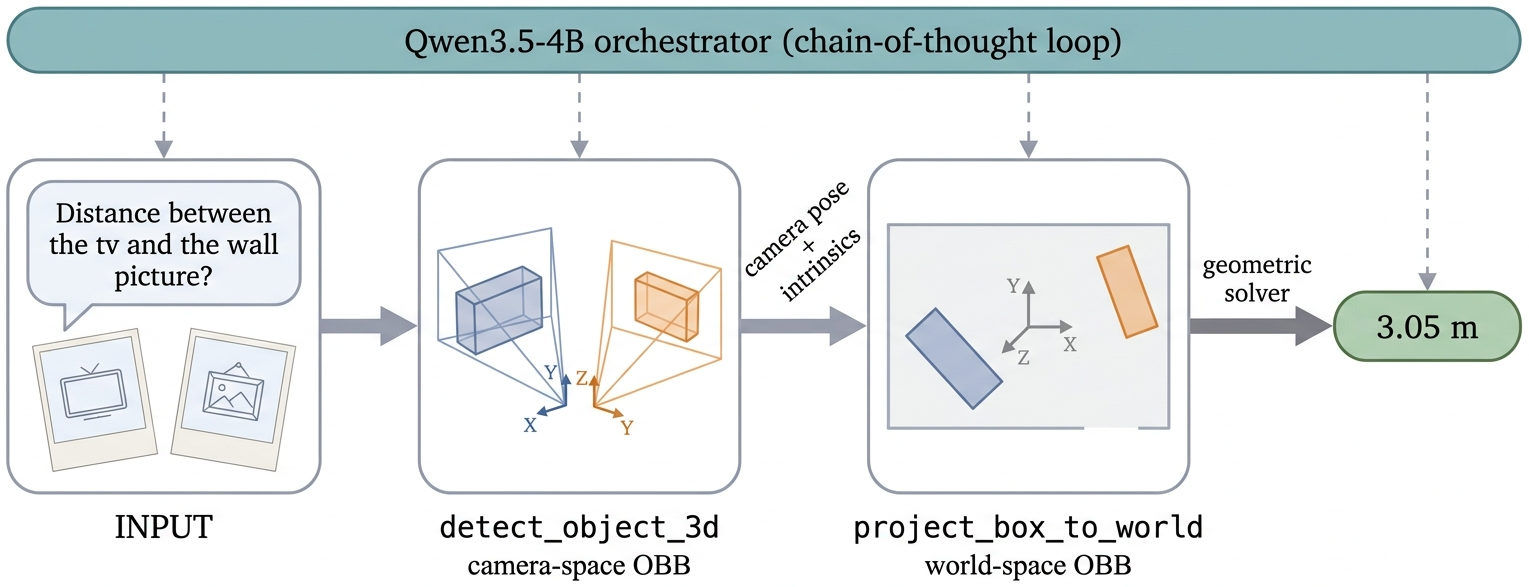}
  \caption{\textbf{Example of tool-augmented metric spatial reasoning.} The VLM orchestrates tools to answer a spatial distance question from video frames: \texttt{detect\_object\_3d} returns an oriented 3D box in the \emph{camera} frame of each object's own best view, \texttt{project\_box\_to\_world} lifts those boxes into a single shared metric world frame using the frames' camera poses, and a deterministic solver returns the surface-to-surface distance.}
  \label{fig:teaser}
\end{figure}

This bottleneck is documented by benchmarks such as VSI-Bench~\cite{vsibench} and its corrected successor ReVSI-Bench~\cite{revsi}. On the ARKitScenes subset, Qwen3.5-4B~\cite{qwen} reaches only $0.46$ Mean Relative Accuracy (MRA) on absolute distance, and smaller models fall further, to $0.21$ (2B) and $0.00$ (0.8B). Simply supplying the missing information does not help: appending the camera intrinsics to the prompt as text lowers MRA to $0.17$. Metric geometry is better encapsulated behind something the model can call than verbalized into its context window.

We present a modular, predictor-agnostic framework (\cref{fig:teaser}) that equips Qwen3.5-4B with geometric tools to detect camera-space 3D boxes, project them into a shared world coordinate system, and compute distances, sizes, and bearing angles, orchestrated through a chain-of-thought (CoT) tool-calling loop. Unlike prior tool-augmented work (\cref{sec:related}), we ask whether a single, off-the-shelf small VLM can orchestrate such tools on its own, and when it fails, whether the fault lies in its tool-calling or in the underlying 3D perception. The perception backend swaps freely behind the same tool interface: a ground-truth (GT) oracle, real monocular 3D detectors (Cube R-CNN~\cite{cubercnn}, OVMono3D~\cite{ovmono3d}, WildDet3D~\cite{wilddet3d}), or a 2D grounding plus metric-depth pipeline (Grounding DINO~\cite{groundingdino} with Depth Pro~\cite{depthpro}).

With a strong detector the tools improve the small model on all four tasks, though only marginally on object size (\cref{sec:exp}). The largest gain is on relative direction, where the unaided model scores $25.9\%$ against a chance rate of $29.3\%$, no better than a blind model that never sees the frames ($28.3\%$); with tools it reaches $73.4\%$. Substituting the GT oracle behind the same tool chain then localizes the residual error: absolute distance climbs to $0.94$ MRA, $0.03$ below the \emph{geometric ceiling}, the score the same tools reach when a script rather than the VLM calls them. This leaves the monocular detector as the binding constraint, and it binds hard. On object size the tools are near-exact on ground-truth boxes ($0.97$), yet the best real detector only just clears the $0.58$ no-tool baseline ($0.61$); on absolute distance every backend but one falls \emph{below} the $0.46$ baseline. Relative direction is the one task limited by the model rather than perception: it caps at $80\%$ against a $100\%$ ceiling, skipping the tools on $23\%$ of questions.

\paragraph{Contributions.}
\begin{itemize}[noitemsep,leftmargin=*,topsep=2pt]
  \item A \textbf{predictor-agnostic tool framework} that lifts camera-space detections into a shared, oriented world frame using the camera poses, letting a small VLM reason metrically across views. The same tool chain runs on monocular 3D detectors or on 2D grounding with metric depth.
  \item An \textbf{orchestration-perception error decomposition} over four ReVSI-Bench tasks. Comparing each detector backend, behind a fixed tool chain, against a ground-truth oracle and a VLM-free geometric ceiling separates the two error sources and attributes the residual error on three tasks to the detector.
  \item An \textbf{analysis of tool-use behavior}. The off-the-shelf VLM already discovers correct tool sequencing without an explicit recipe; what it lacks is the discipline to invoke the tools on questions that appear answerable by inspection, which we identify as the target for fine-tuning.
\end{itemize}

%% file: sec/2_related.tex
\section{Related Work}
\label{sec:related}

\paragraph{Benchmarks for spatial reasoning.}
Early spatial-intelligence evaluations for VLMs focused on 2D relationships (e.g., ``left of'', ``inside'') on datasets like GQA~\cite{gqa}, missing the metric 3D understanding embodied interaction needs. VSI-Bench~\cite{vsibench} and its corrected successor ReVSI-Bench~\cite{revsi} instead probe metric spatial reasoning over multi-view video of indoor scans, and MMSI-Video-Bench~\cite{mmsivideobench} broadens the task suite. MV-RoboBench~\cite{mvrobobench} makes the multi-view setting explicit for robotic manipulation and finds that single-view spatial competence does not reliably transfer to multi-camera scenes. We evaluate on ReVSI-Bench because it provides, per question, the camera poses and intrinsics our tools need plus the ground-truth 3D boxes for an oracle detector; benchmarks without camera calibration are out of reach for this framework.

\paragraph{Baking spatial ability into the model.}
A broad wave of work trains 3D spatial competence into the VLM itself: supervised fine-tuning (MM-Spatial~\cite{mmspatial}, Visual Spatial Tuning~\cite{vst}, SpatialLadder~\cite{spatialladder}, Spatial-MLLM~\cite{spatialmllm}), spatial-reward RL (SpatialThinker~\cite{spatialthinker}), 3D-reconstruction architectures (VLM-3R~\cite{vlm3r}), and scaling studies (Cambrian-S~\cite{cambrians}, Scaling Spatial Intelligence~\cite{scalingspatial}). All change the model's weights; we instead leave it untouched and ask how far a small, in-context VLM gets through tool orchestration alone.

\paragraph{Tool-augmented and agentic VLMs.}
Several works equip VLMs with external tools to bypass unreliable direct reasoning, interleaving reasoning with tool calls as in ReAct~\cite{react} rather than baking the calls into the weights as in Toolformer~\cite{toolformer}: SpaceTools~\cite{spacetools} uses two-phase interactive RL to teach a VLM to compose pointing and depth tools on a single image. Agentic frameworks (Visual Programming~\cite{visprog}, ViperGPT~\cite{vipergpt}) compile queries into executable programs. GCA~\cite{gca} and SpatialPIN~\cite{spatialpin} combine prompting with geometric constraints and 3D priors, and ViSRA~\cite{visra} pairs a VLM with camera-lifted point clouds to build an explicit 3D scene. We instead target multi-view, video-based geometry with a training-free orchestration loop across frames.

%% file: sec/3_method.tex
\section{Method}
\label{sec:method}

\subsection{Dataset}
We evaluate on the ARKitScenes subset of ReVSI-Bench~\cite{revsi} across four question categories: absolute distance (401 questions), object size (423), relative distance (215), and relative direction (290), each question coming with 16 candidate frames sampled from the scene's video. Unlike VSI-Bench~\cite{vsibench}, ReVSI-Bench verifies the queried objects are actually visible across these frames, guaranteeing the question is answerable in the first place. We merge this with the original ARKitScenes camera intrinsics and extrinsics for all 16 frames and the ground-truth 3D boxes of the objects relevant to each question, provided by ReVSI-Bench. Numeric tasks (distance, size) use the benchmark's Mean Relative Accuracy, $\text{MRA} = \tfrac{1}{|\mathcal{C}|}\sum_{\theta\in\mathcal{C}} \mathbf{1}[\,|\hat{x}-x|/x < 1-\theta\,]$ over the confidence thresholds $\mathcal{C}=\{0.5,0.55,\dots,0.95\}$; classification tasks (relative distance, direction) use exact-letter accuracy.

\subsection{Frame Selection}
Rather than handing all 16 frames to the VLM, we select only the frames a question needs: two for a distance comparison, one for object size. For each relevant object we project its ground-truth 3D box corners into every candidate frame using that frame's camera pose and intrinsics, and keep the frame(s) where the projected corners stay furthest inside the image bounds, i.e.\ where the object is most visible. Passing all 16 frames per question is computationally impractical, and the 3D detector fails outright on an object a frame does not clearly show. Letting the VLM pick its own frames proved unreliable, so selection happens up front; it is therefore load-bearing rather than preprocessing, and \cref{sec:conclusion} returns to its reliance on ground-truth boxes.

\subsection{Tool-Use Loop}
At each turn the VLM (Qwen3.5-4B) receives the selected frame(s), the question, and a system prompt listing the available tools, their JSON signatures, a pipeline-specific hint on the order to chain them, and the answer format (\cref{sec:suppl_prompts}). Rather than reading metric quantities off the pixels, it acts as an orchestrator (\cref{fig:teaser}): it reasons in a chain-of-thought (CoT) loop and calls tools one at a time via JSON payloads wrapped in \texttt{\textless{}tool\_call\textgreater{}} tags, passing only object identifiers (e.g., \texttt{"sofa"}). Coordinates and matrices stay cached in the host environment, so the context never fills with numbers and the model never does the arithmetic itself.

\subsection{Toolsets and Pipelines}
A per-task toolset (\cref{tab:pipelines}, \cref{sec:suppl_tools}) exposes the tools the question needs and hints at the order to chain them.

\paragraph{Absolute Distance.}
The baseline \texttt{gdino\_\allowbreak depthpro\_\allowbreak abs\_\allowbreak dist} pipeline detects a 2D box (Grounding DINO~\cite{groundingdino}), estimates depth (Depth Pro~\cite{depthpro}), and computes center-to-center distance. Since ReVSI-Bench ground truth measures surface-to-surface distance, \texttt{bbox\_3d\_abs\_dist} instead detects 3D camera-space OBBs (\texttt{detect\_object\_3d}), lifts them to world space (\texttt{project\_box\_to\_world}), and computes the minimum surface distance (\texttt{calculate\_object\_distance}). The choice of geometry accounts for most of the achievable accuracy: on GT boxes, center-to-center distance has an MAE of $0.644$~m against the benchmark answer, versus $0.077$~m for surface-to-surface.

\paragraph{Object Size.}
Object size is the longest dimension of the oriented 3D extent. The \texttt{bbox\_3d\_size} pipeline detects the camera-space box and extracts its longest side (\texttt{calculate\_object\_size}); no world-frame projection is needed since extent is rotation-invariant. The \texttt{gdino\_depthpro\_size} baseline instead back-projects the 2D box, which measures the object's apparent projection and ignores its extent along the viewing direction.

\paragraph{Relative Distance and Direction.}
Relative distance ranking (\texttt{bbox\_3d\_reldist}) generalizes the absolute distance chain across candidates to find the closest/farthest option. Relative direction tracking (\texttt{bbox\_3d\_reldir}) projects viewer, facing, and target into a unified world space and classifies the egocentric bearing angle $\theta$ into 3-way or 4-way sectors.

\subsection{Predictor-Agnostic Detector Backends}
Swapping the backend behind \texttt{detect\_object\_3d} changes only where the camera-space boxes come from, never the tool-calling structure. A \textbf{GT oracle} isolates orchestration; \textbf{monocular 3D detectors} (Cube R-CNN~\cite{cubercnn}, OVMono3D~\cite{ovmono3d}, WildDet3D~\cite{wilddet3d}) substitute real predictions, cached for reproducibility.

\FloatBarrier

%% file: sec/4_experiments.tex
\section{Experiments}
\label{sec:exp}

We run on the full ReVSI-Bench ARKitScenes splits of \cref{sec:method}, using cached backend predictions for determinism; the orchestrator is Qwen3.5-4B, decoded greedily (\cref{sec:suppl_impl}).

\subsection{Baselines and Model Scale}
\begin{table}[t]
  \centering\footnotesize
  \setlength{\tabcolsep}{5pt}
  \renewcommand{\arraystretch}{0.8}
  \begin{tabular}{@{}lccc@{}}
    \toprule
    \multicolumn{4}{c}{No-tool baselines, absolute distance (MRA $\uparrow$)}\\
    \midrule
    Model & visual & blind & +intrinsics \\
    \midrule
    Qwen3.5-0.8B & 0.00 & 0.00 & 0.02 \\
    Qwen3.5-2B   & 0.21 & 0.00 & 0.03 \\
    Qwen3.5-4B   & \textbf{0.46} & 0.29 & 0.17 \\
    \bottomrule
  \end{tabular}
  \caption{Native (no-tool) VLMs on absolute distance. Only the 4B model is a usable baseline; passing camera intrinsics as text does not help.}
  \label{tab:baselines}
\end{table}

Native (no-tool) VLMs degrade sharply below 4B (\cref{tab:baselines}), leaving Qwen3.5-4B as the only usable baseline; even there \texttt{+intrinsics} costs it $0.29$ MRA.

\subsection{Main Results}
\begin{table}[t]
  \centering\footnotesize
  \setlength{\tabcolsep}{4pt}
  \renewcommand{\arraystretch}{0.8}
  \begin{tabular}{@{}lcc@{}}
    \toprule
    Method & MRA $\uparrow$ & Ceiling $\uparrow$ \\
    \midrule
    \multicolumn{3}{l}{Absolute distance (401 Q)}\\
    No tools (visual) & 0.46 & -- \\
    Tools, 2D + depth & 0.32 & 0.29 \\
    Tools, 3D boxes (OVMono3D) & 0.26 & 0.26 \\
    Tools, 3D boxes (Cube R-CNN) & 0.32 & 0.28 \\
    Tools, 3D boxes (WildDet3D) & \underline{0.74} & \underline{0.74} \\
    Tools, 3D boxes (GT oracle) & \textbf{0.94} & \textbf{0.97} \\
    \midrule
    \multicolumn{3}{l}{Object size (423 Q)}\\
    No tools (visual) & 0.58 & -- \\
    Tools, 2D + depth & 0.34 & 0.34 \\
    Tools, 3D boxes (Cube R-CNN) & 0.34 & 0.34 \\
    Tools, 3D boxes (WildDet3D) & \underline{0.61} & \underline{0.62} \\
    Tools, 3D boxes (GT oracle) & \textbf{0.97} & \textbf{1.00} \\
    \bottomrule
  \end{tabular}
  \caption{Absolute distance and object size results (ReVSI-Bench ARKitScenes); the toolset behind each row is given in \cref{tab:pipelines}. ``Ceiling'' runs the geometric tools directly on the boxes, without a VLM, so it measures what the perception backend allows before any orchestration.}
  \label{tab:main}
\end{table}

Under the GT oracle, our framework reaches $0.94$/$0.97$ VLM MRA against a tool ceiling of $0.97$/$1.00$: orchestration costs $0.03$ MRA, and the rest is perception.

Real detectors bear this out: every backend but one falls below the $0.46$ no-tool baseline. OVMono3D reaches $0.26$, Cube R-CNN $0.32$, and the 2D+depth pipeline $0.32$, the last undone by depth outliers of up to $10\,187$~m. A bad box is worse than no box at all (\cref{sec:suppl_perception}); where a backend misses objects, the model's fallback guess can even push its MRA above its own ceiling (\cref{sec:suppl_impl}). Only WildDet3D pays off, lifting distance from $0.46$ to $0.74$ at full coverage.

Object size is bounded by the detector alone. The tools are near-exact given good boxes ($0.97$ against a $1.00$ ceiling), yet the best real detector barely converts that into a gain: WildDet3D reaches $0.61$ against the $0.58$ baseline, because size reads straight off the box extent and WildDet3D's longest side is a median $14.3\%$ off. Distance tolerates a noisy box, since surface distance is dominated by the separation of the centers; size does not.

\subsection{Multiple-Choice Tasks}
\begin{table}[t]
  \centering\footnotesize
  \setlength{\tabcolsep}{4pt}
  \renewcommand{\arraystretch}{0.8}
  \begin{tabular}{@{}lccc@{}}
    \toprule
    Method & Acc.\ $\uparrow$ & Parse & Ceiling \\
    \midrule
    \multicolumn{4}{l}{Relative distance (215 Q, chance $25.0\%$)}\\
    No tools (blind) & 30.2\% & 100.0\% & -- \\
    No tools (visual) & 39.1\% & 100.0\% & -- \\
    Tools, 3D boxes (WildDet3D) & 67.4\% & 100.0\% & 69.3\% \\
    Tools, 3D boxes (GT oracle) & \textbf{92.1\%} & 100.0\% & 94.9\% \\
    \midrule
    \multicolumn{4}{l}{Relative direction (290 Q, chance $29.3\%$)}\\
    No tools (blind) & 28.3\% & 100.0\% & -- \\
    No tools (visual) & 25.9\% & 100.0\% & -- \\
    Tools, 3D boxes (WildDet3D) & 73.4\% & 98.3\% & 89.0\% \\
    Tools, 3D boxes (GT oracle) & \textbf{80.0\%} & 96.6\% & 100.0\% \\
    \bottomrule
  \end{tabular}
  \caption{Multiple-choice results on ReVSI-Bench. ``Parse'' is the share of runs emitting a usable option letter (\cref{sec:suppl_impl}); unparseable answers score wrong. Both no-tool baselines parse at $100\%$, so their at-or-below-chance relative-direction accuracy is a real reasoning failure and not a formatting artifact.}
  \label{tab:mc_tasks}
\end{table}

\paragraph{Relative Distance.}
Tool calling lifts accuracy from $39.1\%$ (visual) to $67.4\%$ (WildDet3D) and $92.1\%$ (GT), $2.8$ points below the GT ceiling ($94.9\%$), which itself falls short of $100\%$ because the ground truth scores against the closest instance of a name (\cref{sec:conclusion}).

\paragraph{Relative Direction.}
Chance is $29.3\%$ ($149$ three-option and $141$ four-option questions). Both no-tool baselines sit at or below it, and the visual baseline ($25.9\%$) does not beat the blind one ($28.3\%$), so the frames carry no orientation signal the VLM can use. The bearing tool raises accuracy to $80.0\%$ (GT) and $73.4\%$ (WildDet3D). The two ceilings separate the error sources: perception costs $11$ points here ($100.0\to89.0\%$) against $25.6$ on relative distance, since a bearing reads object centers and never their extents (\cref{sec:suppl_reldir}). Orchestration costs more, $15.6$ points below the WildDet3D ceiling, because on $23\%$ of questions the model calls no tool and answers directly (\cref{sec:suppl_qual}).

\subsection{No-Recipe Autonomous Probe}
\begin{table}[t]
  \centering\footnotesize
  \setlength{\tabcolsep}{4pt}
  \renewcommand{\arraystretch}{0.8}
  \begin{tabular}{@{}lcccc@{}}
    \toprule
    & \multicolumn{2}{c}{MRA $\uparrow$} & \multicolumn{2}{c}{Acc.\ $\uparrow$} \\
    \cmidrule(lr){2-3}\cmidrule(lr){4-5}
    Method & Dist. & Size & R.\ dist & R.\ dir \\
    \midrule
    No recipe (generic prompt) & \textbf{0.94} & 0.97 & \textbf{92.5} & 70.0 \\
    \, + few-shot order examples & 0.89 & \textbf{1.00} & 80.0 & 75.0 \\
    \midrule
    Scripted recipe (\cref{tab:main,tab:mc_tasks}) & \textbf{0.94} & 0.97 & 92.1 & \textbf{80.0} \\
    \bottomrule
  \end{tabular}
  \caption{Autonomous probe: every tool exposed at once, no per-task recipe ($160$ questions, $40$ per task, GT backend). Distance and size are scored by MRA, the two relative tasks by accuracy. The scripted-recipe row repeats the GT-oracle results of \cref{tab:main,tab:mc_tasks}, measured on the full splits rather than on this subsample.}
  \label{tab:autonomous}
\end{table}

We expose all tools at once behind a generic prompt that names them but prescribes no order, over $40$ questions from each of the four tasks, with the most generous of the scripted budgets ($20$ steps, $2048$ tokens; \cref{sec:suppl_impl}).

Reaching a working autonomous loop meant fixing the tools, not the prompt. The original error strings were terse (``No 3D box found for [tv, sink]''), so a mis-ordered call told the model nothing about what to do next and it simply retried. Rewriting each error to name the missing step removes most of that looping: in an $80$-question ablation it cuts tool errors from $138$ to $36$ and raises distance from $0.58$ to $0.88$ MRA (\cref{sec:suppl_auto}); all runs here use the rewritten errors. With that feedback the recipe becomes unnecessary (\cref{tab:autonomous}): the generic prompt reaches $0.94$ and $0.97$ MRA on distance and size and $92.5\%$ on relative distance, matching the scripted pipeline on all three. Few-shot ordering examples do not improve on it: they gain on size and relative direction but cost $0.05$ MRA on distance and $12.5$ points of relative-distance accuracy. Relative direction again lags at $70.0\%$, where the model skips the tools. Routing capability is already present; the discipline to invoke the tools is not.

%% file: sec/5_conclusion.tex
\section{Conclusion and Next Steps}
\label{sec:conclusion}

\paragraph{Conclusion.}
Encapsulating 3D geometry behind modular tools lets a small VLM perform multi-view metric spatial reasoning it cannot do from pixels alone. Comparing each detector against a ground-truth oracle attributes the residual error to monocular 3D perception: orchestration costs only $0.03$ MRA, while a weak detector is worse than no tool at all.

\paragraph{Limitations.}
Our framework selects one ``best frame'' per object \emph{using the ground-truth boxes}, and never triangulates across views. The no-tool baselines also see all 16 frames while the tool runs see only the selected one or two, so part of the gain may come from framing rather than from the tools; the GT-oracle numbers are invariant to the frame chosen. For relative distance we map each name to a single predicted box ($5.1\%$ ceiling penalty). All three detectors are trained on Omni3D~\cite{cubercnn}, which contains the ARKitScenes training split; our $161$ scenes come from the validation fold, so none is seen in training, but perception stays in-domain.

\paragraph{Next steps.}
Frame selection is the first oracle to remove: the visibility score should become a tool the model calls, scored by the open-vocabulary 2D detector we already expose. A feed-forward pose predictor~\cite{vggt} would remove the second oracle, letting the chain run on uncalibrated video. Beyond that: better box extents, multi-frame triangulation, instance-aware routing, and fine-tuning.

%% file: sec/6_acknowledgments.tex
\section*{Acknowledgments}
We thank our supervisor, Bartlomiej Baranowski, for his guidance and feedback throughout this project.

%% file: sec/X_suppl.tex
\clearpage
\setcounter{page}{1}
\maketitlesupplementary

% ---------------------------------------------------------------------------
% OPTIONAL appendix: EXTRA DETAIL + VISUALS. The main report stands on its own;
% nothing here is required to understand or evaluate the work.
%
% Restart every counter so the supplement is numbered independently of the main
% body: sections become A, B, C ... and floats become Figure A1, Table A1,
% Listing A1 (instead of continuing on from the main text as 6, 7, Figure 2 ...).
% ---------------------------------------------------------------------------
\setcounter{section}{0}
\renewcommand{\thesection}{\Alph{section}}
\setcounter{figure}{0}
\renewcommand{\thefigure}{A\arabic{figure}}
\setcounter{table}{0}
\renewcommand{\thetable}{A\arabic{table}}
\setcounter{lstlisting}{0}
\renewcommand{\thelstlisting}{A\arabic{lstlisting}}

% Restarting the section and page counters also restarts the PDF anchor names
% hyperref mints from them (section.1, page.1, ...), which already exist in the
% main body -> "destination with the same identifier" warnings and dead links.
% Give the supplement's anchors their own namespace. Page anchors are only needed
% by pagebackref, which resolves citations in the main body, so dropping them here
% is safe (the supplement cites nothing).
\renewcommand{\theHsection}{suppl.\Alph{section}}
\hypersetup{pageanchor=false}

% The supplement places its figures inline ([H], see Sec. E). Under the two-column
% default (\flushbottom) LaTeX stretches the glue between them to equalise column
% heights, which opened a gap in the middle of a column whenever the next figure did
% not fit. \raggedbottom lets that leftover space collect at the column bottom.
\setlength{\intextsep}{10pt plus 2pt minus 2pt}
\raggedbottom

\section{Implementation Details}
\label{sec:suppl_impl}

\paragraph{Orchestrator and decoding.}
Every tool-use run drives the same orchestrator, Qwen3.5-4B in \texttt{bfloat16},
with greedy decoding (\texttt{do\_sample=False}); no sampling temperature is
involved, so a run reproduces its own trace. Each selected frame enters the VLM
at a budget of $16$--$144$ visual tokens ($\approx\!112^2$--$336^2$~px). The
generation budget is $1024$ new tokens per turn on the numeric tasks and $2048$
on the multiple-choice and autonomous runs, so that a long chain-of-thought
cannot truncate mid-tool call.

\paragraph{Step caps.}
The tool-use loop is capped per task, always with headroom over the minimal chain
(\cref{sec:suppl_steps}): $5$ steps for object size (minimal $2$), $10$ for
absolute distance (minimal $5$), $14$ for relative direction (minimal $7$), and
$20$ for relative distance (minimal $14$) and the autonomous probe. A run that
hits the cap is scored on whatever answer it has produced, i.e.\ as a miss unless
it already emitted one.

\paragraph{Answer parsing.}
Numeric answers are the last number in the final message; an unparseable answer
scores $\mathrm{MRA}=0$ (a miss) but contributes no metre value, so it is excluded
from the error statistics of \cref{fig:absdist-err}. Multiple-choice answers are
the option letter, or the uniquely matching option name if no bare letter is
emitted; the resulting parse rates are the \emph{Parse} column of
\cref{tab:mc_tasks}.

\paragraph{Cached backends and coverage.}
\texttt{detect\_object\_3d} reads pre-computed camera-space boxes, so every
backend is deterministic and the tool-calling structure is identical across them
(\cref{sec:method}). A backend that has no box for a queried name returns an
error string instead, and the model then answers from the frames alone.
\Cref{tab:coverage} gives the resulting coverage on the absolute-distance split.
This is the mechanism behind the two inversions in the main results. Cube R-CNN
misses $8.2\%$ of objects, and on the $15\%$ of questions where a box is missing
the VLM falls back to guessing, which beats the zero MRA the tool ceiling
records for a missing box, so its VLM MRA ($0.32$) \emph{exceeds} its own ceiling
($0.28$); the 2D grounding front-end inverts for the same reason ($0.32$ against
a $0.29$ ceiling).

\begin{table}[H]
  \centering\footnotesize
  \setlength{\tabcolsep}{5pt}
  \begin{tabular}{@{}lcc@{}}
    \toprule
    Backend & Objects found & Questions fully covered \\
    \midrule
    GT Oracle  & 100.0\% (802/802) & 100.0\% (401/401) \\
    OVMono3D   & 100.0\% (802/802) & 100.0\% (401/401) \\
    Cube R-CNN & 91.8\% (736/802)  & 85.0\% (341/401) \\
    WildDet3D  & 100.0\% (802/802) & 100.0\% (401/401) \\
    Grounding DINO (2D) & 93.6\% (751/802) & 87.8\% (352/401) \\
    \bottomrule
  \end{tabular}
  \caption{Detector coverage on the absolute-distance split (401 questions, 802 queried object instances). \emph{Objects found}: the backend returns a box for the queried name. \emph{Questions fully covered}: boxes exist for \emph{both} objects, i.e.\ the tool chain can run end-to-end. Cube R-CNN and the 2D grounding front-end miss objects, which is why their VLM MRA can exceed their own ceiling; OVMono3D is fully covered yet still the weakest backend (\cref{tab:main}), so coverage and accuracy are independent failure modes.}
  \label{tab:coverage}
\end{table}

\FloatBarrier
\section{System Prompts and Toolset Hints}
\label{sec:suppl_prompts}

Every run assembles its system prompt from a single template
(\cref{lst:scaffold}): a fixed preamble, the JSON schemas of the tools exposed
for that pipeline, the pipeline-specific \emph{hint} (the recipe), a
one-tool-at-a-time calling convention, and the task-specific answer format. Only
the tool subset, the task/answer-format strings, and the hint change between
pipelines; the scaffold is identical. \Cref{tab:promptfields} lists the
per-pipeline task and answer-format strings, and
\cref{lst:hint-bbox3d,lst:hint-auto} give two hints verbatim. The other four (2D+depth
distance, object size, relative distance, relative direction) follow the same
detect $\to$ project $\to$ measure shape and are omitted for space; \cref{tab:pipelines}
gives each pipeline's tool chain.

% [H] pins the table exactly here, ahead of the listings. As a real float it
% would drift: to the column top (above the section heading), or to the column
% bottom (splitting the listing that follows it).
\begin{table}[H]
  \centering\footnotesize
  \setlength{\tabcolsep}{4pt}
  \begin{tabular}{@{}p{3.7cm}p{3.8cm}@{}}
    \toprule
    Pipeline & \texttt{task} / \texttt{answer\_format} \\
    \midrule
    \texttt{bbox\_3d\_abs\_dist} \newline \texttt{gdino\_depthpro\_abs\_dist} &
      distances between objects \newline \emph{a single number in metres, e.g.: 1.5} \\
    \addlinespace
    \texttt{bbox\_3d\_size} \newline \texttt{gdino\_depthpro\_size} &
      the size (longest dimension) of objects \newline \emph{a single number in centimetres, e.g.: 120} \\
    \addlinespace
    \texttt{bbox\_3d\_reldist} &
      which of several candidate objects is closest or farthest to a reference object \newline \emph{the letter (A, B, C, or D)} \\
    \addlinespace
    \texttt{bbox\_3d\_reldir} &
      the egocentric direction (left/right/back, or a front/back-left/right quadrant) of a target from a viewer \newline \emph{the letter (A, B, C, or D)} \\
    \addlinespace
    \texttt{autonomous\_3d} &
      spatial questions about objects (all four tasks mixed) \newline \emph{the exact format the question asks for} \\
    \bottomrule
  \end{tabular}
  \caption{Per-pipeline \texttt{task} and \texttt{answer\_format} strings substituted into the scaffold of \cref{lst:scaffold}. These reuse the same tool interface across numeric (MRA) and multiple-choice tasks.}
  \label{tab:promptfields}
\end{table}

\begin{lstlisting}[style=prompt,caption={Prompt scaffold (\texttt{build\_system\_prompt}). \texttt{\{tools\}} is the pretty-printed JSON of the exposed tool schemas; \texttt{\{hint\}}, \texttt{\{task\}} and \texttt{\{answer\_format\}} are filled per pipeline.},label={lst:scaffold}]
You are a precise spatial measurement assistant.
You answer questions about {task} by calling tools.

Available tools:
{tools}

{hint}

Call ONE tool at a time and wait for its result before
calling the next tool.
To call a tool, use this exact format:
<tool_call>{"name": "tool_name",
 "arguments": {"arg": "value"}}</tool_call>

When you have computed the answer, output it as
{answer_format}
\end{lstlisting}

\begin{lstlisting}[style=prompt,caption={Hint for \texttt{bbox\_3d\_abs\_dist} (absolute distance). The recipe enforces the camera$\rightarrow$world lift before any metric comparison.},label={lst:hint-bbox3d}]
detect_object_3d returns a box in CAMERA space,
measured from that object's own camera, so two
such boxes cannot be compared until each is
lifted to world space.
Every tool takes ONLY object names -- never copy
coordinates between calls.
Follow this exact sequence; do not repeat a step:
  1. detect_object_3d for the first object
       -> camera-space box
  2. project_box_to_world for the first object
       -> world box
  3. detect_object_3d for the second object
       -> camera-space box
  4. project_box_to_world for the second object
  5. calculate_object_distance (two names)
       -> distance in metres
Call detect_object_3d and project_box_to_world
exactly once per object, then output the number.
\end{lstlisting}

\begin{lstlisting}[style=prompt,caption={Hint for \texttt{autonomous\_3d} (no-recipe probe, \cref{sec:exp}). The full 5-tool union with no step sequence: the model must route and order the tools itself.},label={lst:hint-auto}]
You have tools to detect an object's 3D box,
lift a box into world coordinates, measure an
object's size, measure the distance between two
objects, and compute the egocentric direction
of one object from a viewer facing another.
Decide which tools to call, and in what order,
to answer THIS question -- not every tool is
needed for every question. First read the
question and identify what it asks: one
object's size, the distance between two
objects, which listed object is closest or
farthest to a reference, or the direction of
an object relative to a viewer. Use the tools
to compute the answer rather than guessing
from the images. Answer in the exact format
the question asks for.
\end{lstlisting}

\section{Tool Library and Geometric Logic}
\label{sec:suppl_tools}

All metric computation lives inside the tools; the orchestrator only passes
object \emph{names}, and each tool reads the cached geometry for that name from
the execution context. This section gives the exact math behind the four
geometric solvers, plus how \texttt{detect\_object\_3d} turns pixels into a
metric box in the first place. Let a detected oriented box be
$(\mathbf{c}, R, \mathbf{e})$ with center $\mathbf{c}\in\mathbb{R}^3$, rotation
$R\in \mathrm{SO}(3)$ (columns are the box axes), and full extent
$\mathbf{e}\in\mathbb{R}^3_{>0}$.

\paragraph{Camera-space detection (\texttt{detect\_object\_3d}).}
Unlike the solvers below, this tool is a predictor-agnostic wrapper around a
monocular 3D detector (Cube R-CNN, OVMono3D, WildDet3D) or the GT oracle. The
real detectors need the frame's camera \emph{intrinsics}: we rescale and
re-orient the native sensor matrix to the upright display frame the model
actually sees, $K_{\text{eff}} = \bigl[\begin{smallmatrix}f_x&0&c_x\\0&f_y&c_y\\0&0&1\end{smallmatrix}\bigr]$
(\texttt{compute\_keff}), and pass $K_{\text{eff}}$ into the model's forward
pass so it can convert its 2D detection and relative depth into a metric
camera-space box. This is the only place intrinsics enter the pipeline;
downstream, \texttt{project\_box\_to\_world} uses only extrinsics.

\paragraph{World projection (\texttt{project\_box\_to\_world}).}
\texttt{detect\_object\_3d} returns a box in the \emph{camera} frame of that
object's best frame. Lifting it to world space needs only that frame's
\emph{extrinsics}: inverting the stored world-to-camera pose
$(R_{w\to c}, \mathbf{t}_{w\to c})$ gives the camera-to-world rigid transform
$(R_{c\to w}, \mathbf{t}_{c\to w})\in\mathrm{SE}(3)$, and
\begin{equation}
  \mathbf{c}_w = R_{c\to w}\,\mathbf{c}_c + \mathbf{t}_{c\to w},\quad
  R_w = R_{c\to w} R_c,\quad
  \mathbf{e}_w = \mathbf{e}_c ,
\end{equation}
i.e.\ the extent is invariant (the lift is rigid). This is what makes two boxes
from \emph{different} cameras comparable; without it a distance between them is
meaningless. No pixel is back-projected here, so the camera \emph{intrinsics}
play no role in this step.

\paragraph{Surface-to-surface distance (\texttt{calculate\_object\_distance}).}
ReVSI ground truth measures the distance between the nearest \emph{surfaces},
not centers. We approximate it by subtracting each oriented box's half-extent,
projected onto the center-separation direction, from the center distance. With
$\hat{\mathbf{d}} = (\mathbf{c}_b-\mathbf{c}_a)/\lVert \mathbf{c}_b-\mathbf{c}_a\rVert$,
\begin{align}
  \mathrm{proj}(\text{box}) &= \bigl|R^{\!\top}\hat{\mathbf{d}}\bigr| \cdot \tfrac{1}{2}\mathbf{e}, \\
  d_{\text{surf}} &= \max\bigl(0,\; \lVert \mathbf{c}_b-\mathbf{c}_a\rVert \notag \\
                  &\qquad\; - \mathrm{proj}(\text{box}_a) - \mathrm{proj}(\text{box}_b)\bigr).
\end{align}
On GT boxes this reduces MAE against the benchmark from $0.644$~m
(center-to-center) to $0.077$~m ($71\%$ of samples within $0.1$~m).

\paragraph{Object size (\texttt{calculate\_object\_size}).}
Size is the longest side of the oriented box, in centimeters:
$\text{size}_{\text{cm}} = 100\cdot\max_i e_i$. Because the extent is
rigid-invariant (above), no world lift is needed and the size pipeline is the
minimal two-step chain \texttt{detect\_object\_3d}$\to$\texttt{calculate\_object\_size}.

\paragraph{Egocentric bearing (\texttt{relative\_direction}).}
Given three world centers projected to the ground plane (heights dropped),
namely viewer $\mathbf{a}$, facing $\mathbf{b}$ and target $\mathbf{c}$, the facing
and target vectors are
\begin{equation}
  \mathbf{f} = \begin{cases}\mathbf{b}-\mathbf{a} & \text{mode }=\text{toward}\\ \mathbf{a}-\mathbf{b} & \text{mode }=\text{away}\end{cases},\qquad
  \mathbf{t} = \mathbf{c}-\mathbf{a},
\end{equation}
and the signed bearing (positive $\Rightarrow$ target on the \emph{left}) is
\begin{equation}
  \theta = \operatorname{atan2}\!\bigl(f_x t_y - f_y t_x,\; \mathbf{f}\cdot\mathbf{t}\bigr).
\end{equation}
It is classified into the answer's direction words. 3-way:
\textsc{back} if $|\theta|\ge 135^\circ$, else \textsc{left}/\textsc{right} by
sign. 4-way: \textsc{front} if $|\theta|<90^\circ$ else \textsc{back},
combined with left/right. This bearing reproduces the ReVSI ground-truth letter
on all $290/290$ relative-direction samples, so the residual error on this task
is orchestration, not geometry.

\Cref{tab:toolregistry} summarizes the full tool registry and which pipelines
expose each tool.

% [H] keeps the registry inside Sec. C; as a float it drifted past the Sec. D
% heading onto a later page.
\begin{table}[H]
  \centering\scriptsize
  \setlength{\tabcolsep}{4pt}
  \begin{tabular}{@{}p{3.6cm}p{4.05cm}@{}}
    \toprule
    Tool & Role (in/out) \\
    \midrule
    \texttt{detect\_object\_3d} & name $\to$ camera-space OBB (predictor-agnostic backend) \\
    \texttt{project\_box\_to\_world} & name $\to$ world-space OBB (pose lift) \\
    \texttt{calculate\_object\_distance} & two names $\to$ surface distance (m) \\
    \texttt{calculate\_object\_size} & name $\to$ longest dim (cm) \\
    \texttt{relative\_direction} & three names $+$ mode $\to$ 3-/4-way label \\
    \midrule
    \texttt{detect\_object} & name $\to$ 2D bbox (Grounding DINO) \\
    \texttt{estimate\_depth} & bbox $\to$ metric depth (Depth Pro) \\
    \texttt{project\_to\_world} & bbox$+$depth $\to$ 3D world point \\
    \texttt{calculate\_distance} & two points $\to$ center distance (m) \\
    \texttt{estimate\_size\_from\_bbox} & bbox$+$depth $\to$ apparent size (cm) \\
    \bottomrule
  \end{tabular}
  \caption{Tool registry. Top block: 3D OBB tools (used by the \texttt{bbox\_3d\_*} and \texttt{autonomous\_3d} pipelines). Bottom block: the 2D + depth contrast tools (\texttt{gdino\_depthpro\_*}). Every tool takes object \emph{names}; geometry is cached host-side.}
  \label{tab:toolregistry}
\end{table}

\Cref{tab:pipelines} groups these tools into the four core pipelines (toolsets) referenced from \cref{sec:method}, one per task, plus the 2D + depth baselines for absolute distance and object size.

\begin{table*}[t]
  \centering\footnotesize
  \setlength{\tabcolsep}{4pt}
  \renewcommand{\arraystretch}{0.8}
  \begin{tabular}{@{}lllcc@{}}
    \toprule
    Task & Pipeline (Toolset) & Geometry Representation & Min. Steps & Median Steps \\
    \midrule
    Abs. Distance & \texttt{gdino\_depthpro\_abs\_dist} & 2D bbox + Depth (center-to-center) & 7 & 7 \\
    Abs. Distance & \texttt{bbox\_3d\_abs\_dist} & 3D OBB (surface-to-surface) & 5 & 5 \\
    Object Size   & \texttt{gdino\_depthpro\_size} & Apparent 2D projection & 3 & 3 \\
    Object Size   & \texttt{bbox\_3d\_size} & 3D OBB extent (rigid invariant) & 2 & 2 \\
    Rel. Distance & \texttt{bbox\_3d\_reldist} & 3D OBB (nearest-instance surface) & 14 & 16 \\
    Rel. Direction & \texttt{bbox\_3d\_reldir} & 3D OBB (egocentric bearing angle) & 7 & 8 \\
    \bottomrule
  \end{tabular}
  \caption{The four core toolsets, plus the 2D + depth baselines. Direct 3D box pipelines compute surface-to-surface relationships and rigid invariants. Median Steps (\cref{sec:suppl_steps}; GT backend for the 3D pipelines, real detector for the 2D) equals Min. Steps on every numeric pipeline and exceeds it only on the two multiple-choice tasks, a benign overhead: a premature bearing call that errors and is retried (rel.\ direction), and redundant distance calls (rel.\ distance).}
  \label{tab:pipelines}
\end{table*}

% No \FloatBarrier here: it would hold Table A4 (a full-width [t] float that cannot
% fit in the remainder of the current page) as pending and block the following text
% from backfilling that page, leaving it two-thirds empty. Letting the table float to
% the next column-top instead lets Sec. D's opening flow up and fill the page.
\section{Qualitative Agent Traces}
\label{sec:suppl_qual}

\Cref{lst:trace-success} shows a complete successful absolute-distance trace
(tv $\leftrightarrow$ wall picture): the model follows the recipe exactly
(detect, lift, detect, lift, measure), and the surface-distance tool returns
$3.05$~m against a $3.20$~m ground truth (MRA $1.00$), while the naive
center-to-center distance would have been $3.34$~m.
\Cref{lst:trace-fail} shows the dominant relative-direction failure mode: the
model calls \emph{zero} tools, reasons directly from the frames, and guesses the
wrong quadrant, which is the ``discipline'' gap analyzed in \cref{sec:exp}.

\begin{lstlisting}[style=prompt,caption={Success trace (\texttt{bbox\_3d\_abs\_dist}, GT backend). CoT abbreviated; tool calls and returns verbatim. GT $3.2$~m, prediction $3.05$~m (MRA $1.00$).},label={lst:trace-success}]
Q: Measuring from the closest point of each
   object, what is the direct distance between
   the tv and the wall picture?

[1] CoT: follow the sequence; start with the tv.
    -> detect_object_3d{object_name: "tv"}
    <- {center_cam:[-1.17,0.36,1.74],
        extent:[0.04,0.89,0.51]}
[2] CoT: lift the tv box to world space.
    -> project_box_to_world{object_name: "tv"}
    <- {world_center:[-2.96,1.32,0.78],
        extent:[0.04,0.89,0.51]}
[3] -> detect_object_3d{object_name:"wall picture"}
    <- {center_cam:[0.87,-0.22,1.43],
        extent:[0.77,0.05,0.80]}
[4] -> project_box_to_world{object_name:
                            "wall picture"}
    <- {world_center:[0.36,0.94,0.79],
        extent:[0.77,0.05,0.80]}
[5] CoT: both in world space; measure the distance.
    -> calculate_object_distance{
         object_a:"tv", object_b:"wall picture"}
    <- {distance_m: 3.05, center_distance_m: 3.34}
[final] Answer: 3.05
\end{lstlisting}

\begin{lstlisting}[style=prompt,caption={Failure trace (\texttt{bbox\_3d\_reldir}, GT backend): a zero-tool guess. The bearing tool would have returned \textsc{left} (A); the model answered \textsc{right} (B) from the frames alone.},label={lst:trace-fail}]
Q: If I am standing by the wall picture and
   facing the radiator, is the dresser to my
   left, right, or back?
   Options: A. left  B. right  C. back  (GT: A)

[final] CoT (no tool call): "wall picture
   visible above the bed headboard; radiator on
   the far wall; the dresser is off to the
   side ... it should be to my right."
    Answer: B   (WRONG -- 0 tools called)
\end{lstlisting}

% Perception figures and quantitative plots share one section, and every section
% from here on ends in a \FloatBarrier (placeins). As two figure-only sections of
% one paragraph each, neither could fill a page, so their [t] floats queued up and
% landed one to two pages later, under a section that did not discuss them.
\section{Perception and Quantitative Visualizations}
\label{sec:suppl_perception}
\label{sec:suppl_plots}

\begin{figure}[H]
  \centering
  \includegraphics[width=\linewidth]{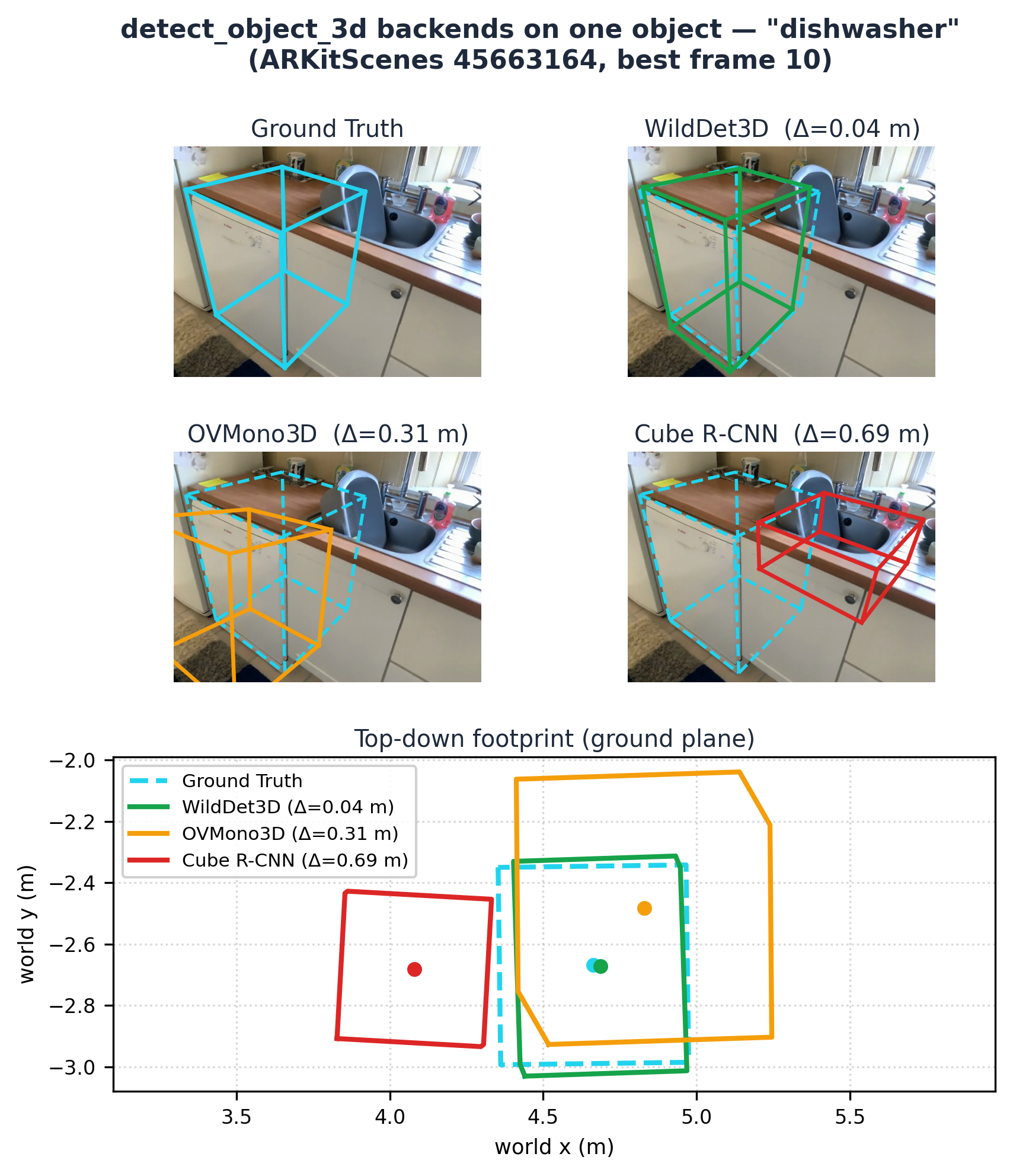}
  \caption{Detector backends behind \texttt{detect\_object\_3d} on one object (dishwasher, ARKitScenes 45663164, best frame 10). \emph{Top four panels:} each backend's predicted 3D box (solid) with the ground-truth box (dashed cyan) reprojected onto the best frame; $\Delta$ is the world-space center error. \emph{Bottom:} the same boxes as top-down ground-plane footprints. WildDet3D ($\Delta{=}0.04$~m) nearly coincides with GT, OVMono3D ($0.31$~m) is oversized and offset, and Cube R-CNN ($0.69$~m) drifts onto the sink. The tool-calling chain is identical across backends; only the box origin changes, so this gap is the perception bottleneck of \cref{sec:exp}.}
  \label{fig:backends}
\end{figure}

\begin{figure}[H]
  \centering
  \includegraphics[width=\linewidth]{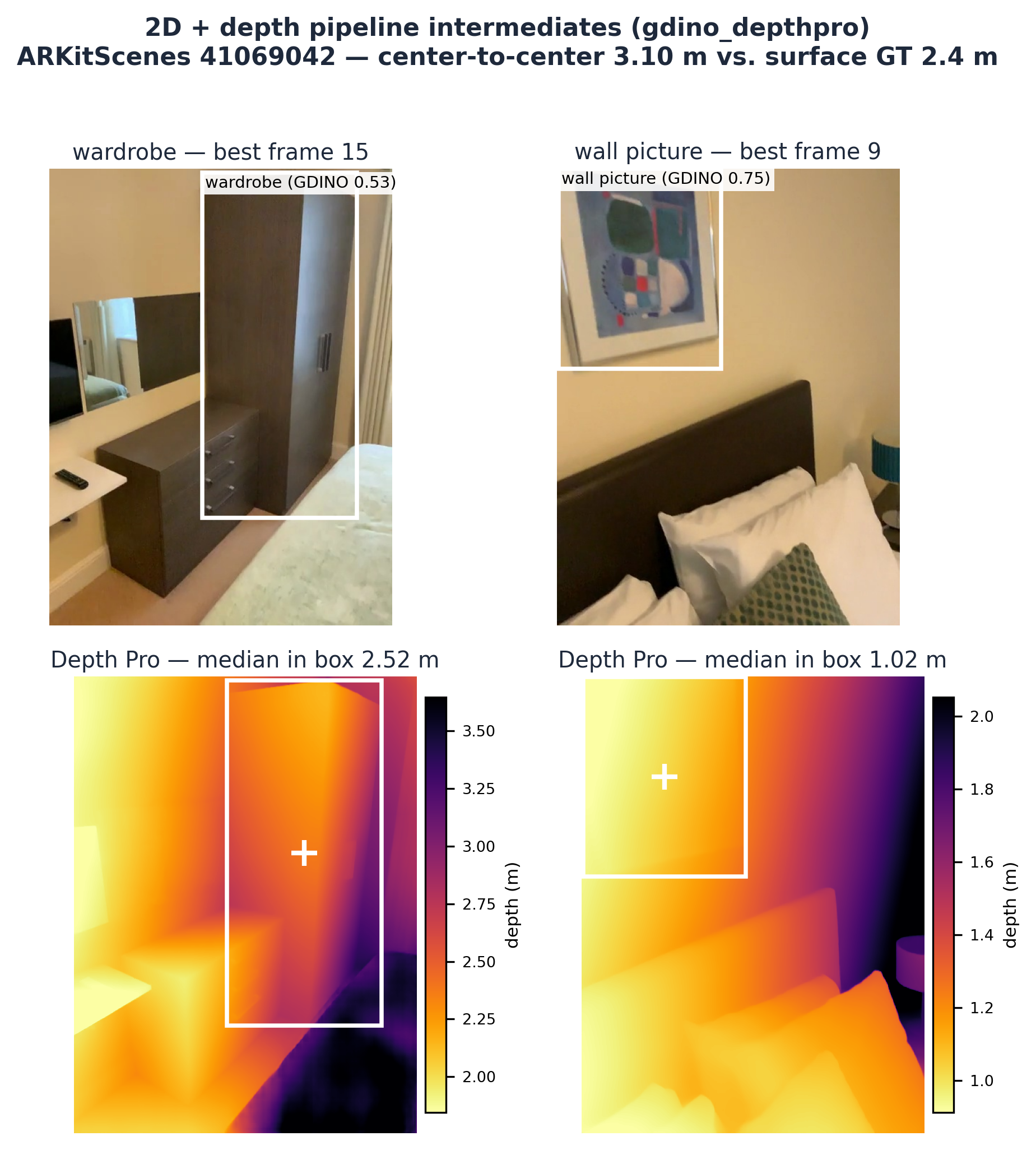}
  \caption{Intermediate signals of the 2D + depth pipeline (\texttt{gdino\_depthpro\_abs\_dist}) on a wardrobe/wall-picture distance query (ARKitScenes 41069042). \emph{Top:} each object's best frame with its Grounding DINO 2D box. \emph{Bottom:} the metric depth map Depth Pro predicts for that frame (bright $=$ near, each map scaled to its own range), with the same box and the sampled center pixel (``$+$''); the pipeline keeps only the median depth inside the box ($2.52$~m for the wardrobe, $1.02$~m for the wall picture). Back-projecting the two box centers yields a center-to-center distance ($3.10$~m) that overshoots the surface-to-surface ground truth ($2.4$~m), the systematic bias the OBB pipeline corrects. Depth Pro also reads the bedspread in the lower right as \emph{far} (dark) although it is the nearest surface: monocular depth degrades on large textureless regions, the weakness behind the error tail in \cref{fig:absdist-err}.}
  \label{fig:perception}
\end{figure}

\paragraph{Perception.}
\Cref{fig:backends} compares the camera-space 3D boxes produced by each
\texttt{detect\_object\_3d} backend on the same object, making the perception
bottleneck of \cref{sec:exp} visible: the OVMono3D box is oversized and offset
and the Cube R-CNN box drifts onto a neighboring object, while WildDet3D closely
tracks the GT oracle.
\Cref{fig:perception} shows the intermediate 2D + depth signals of the
\texttt{gdino\_depthpro\_abs\_dist} pipeline.

\begin{figure}[H]
  \centering
  \includegraphics[width=0.95\linewidth]{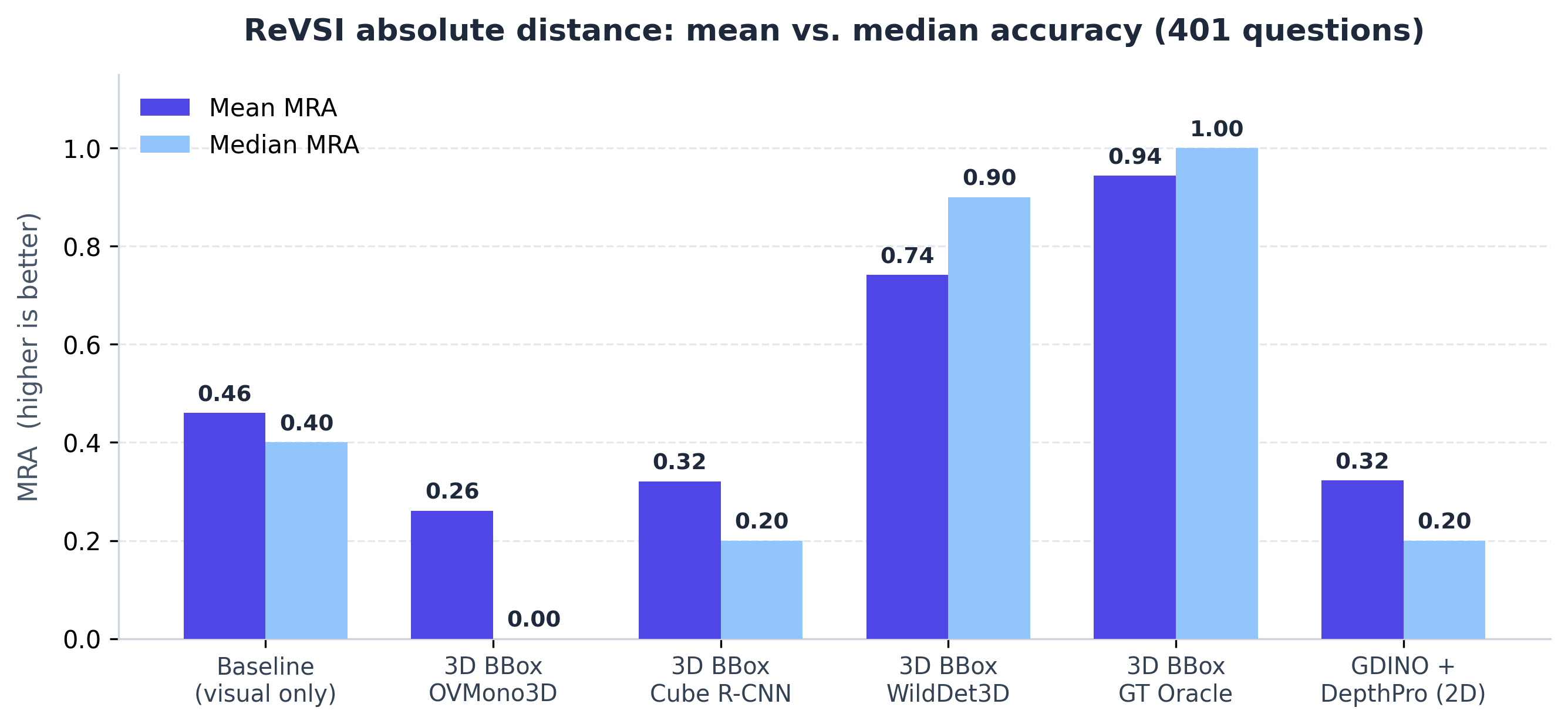}
  \caption{Mean vs.\ median MRA across every absolute-distance pipeline (401 questions); the gap between the bars measures tail sensitivity. Only WildDet3D beats the no-tool baseline: OVMono3D ($0.26$), Cube R-CNN ($0.32$) and the 2D+depth pipeline ($0.32$) all land below it ($0.46$), so a poor box is worse than no box. OVMono3D's median is $0.00$, i.e.\ more than half its answers are off by over $50\%$. WildDet3D (mean $0.74$, median $0.90$) approaches the GT oracle ($0.94$/$1.00$).}
  \label{fig:absdist-acc}
\end{figure}

\begin{figure}[H]
  \centering
  \includegraphics[width=0.95\linewidth]{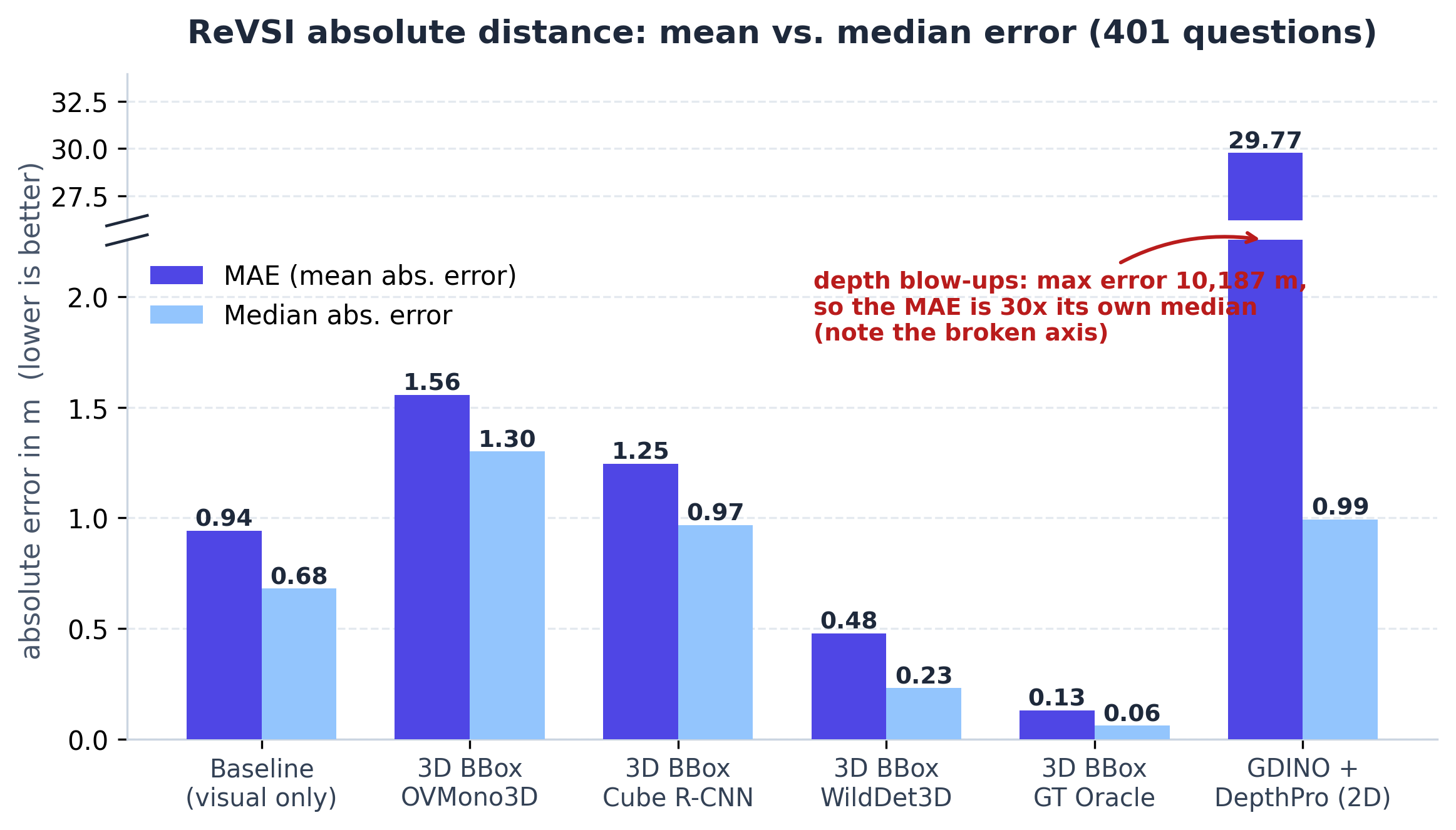}
  \caption{The same pipelines in metric \emph{error}, which corroborates the MRA ranking. The weak 3D backends are worse than the no-tool baseline (MAE $1.56$~m for OVMono3D and $1.25$~m for Cube R-CNN, against $0.94$~m), while WildDet3D ($0.48$~m) and the GT oracle ($0.13$~m) sit far below it. Every OBB pipeline stays bounded (max error $\le 6.2$~m), so their means are trustworthy. The 2D+depth pipeline is the exception: monocular depth blows up on boundary boxes to a worst error of $10\,187$~m, leaving an MAE ($29.8$~m) thirty times its own median ($0.99$~m).}
  \label{fig:absdist-err}
\end{figure}

\begin{figure}[H]
  \centering
  \includegraphics[width=0.95\linewidth]{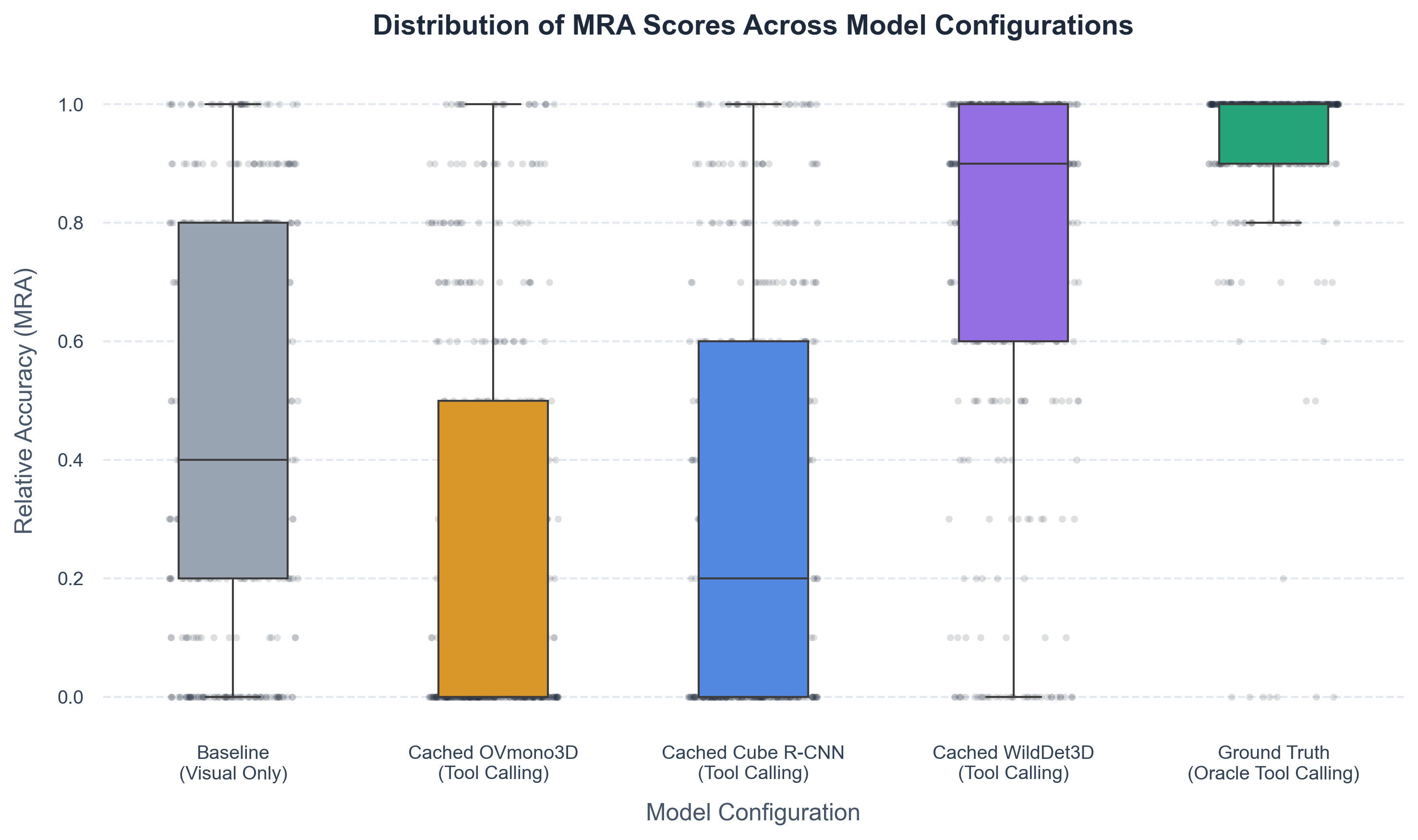}
  \caption{Per-sample MRA distributions on absolute distance (401 questions) for the no-tool visual baseline and each \texttt{detect\_object\_3d} backend. The GT oracle concentrates at $1.0$ (median $1.00$) and WildDet3D tracks it (median $0.90$), whereas OVMono3D and Cube R-CNN pile up at zero: their medians ($0.00$ and $0.20$) sit \emph{below} the no-tool baseline's ($0.40$), so over half of OVMono3D's answers are more than $50\%$ off. Tool augmentation only pays off once the detector is strong enough.}
  \label{fig:mradist}
\end{figure}

\begin{figure}[H]
  \centering
  \includegraphics[width=\linewidth]{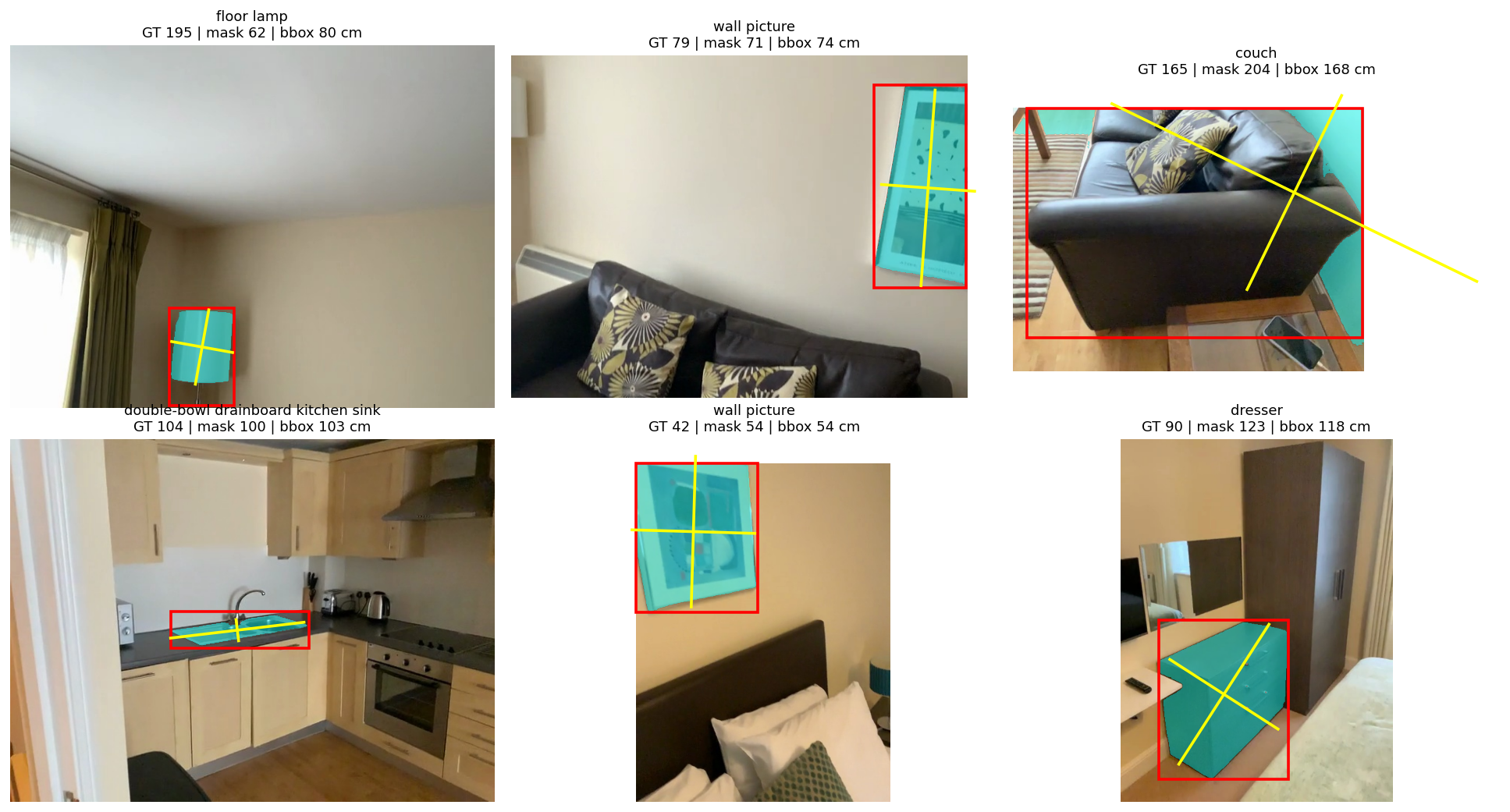}
  \caption{Object size from a 2D box (\texttt{gdino\_depthpro\_size}) on six ReVSI objects: the Grounding DINO box (red) and mask (cyan) with the longest image-plane axis (yellow). Panel titles give GT and both back-projected estimates in cm, mask-based and box-based (\texttt{bbox}, the reported variant). Back-projection sees only the image-plane dimensions, so it is accurate when the longest axis lies in that plane (kitchen sink $104\to103$, couch $165\to168$) and fails in two ways: the box covers only part of the object (floor lamp $195\to80$: only the shade is grounded), or it lands on the wrong surface (dresser $90\to118$). This variance, not a constant bias, caps the 2D size pipeline at $0.34$ MRA.}
  \label{fig:size2d}
\end{figure}

\paragraph{Quantitative plots.}
\Cref{fig:absdist-acc} and \cref{fig:absdist-err} give the same $401$ absolute-distance
questions in accuracy and in metric error; together they are the depth blow-up
argument of \cref{sec:exp}. The 2D + depth baseline has a small median error but a
heavy tail that inflates the MAE, whereas the OBB pipelines are stable.
\Cref{fig:mradist} shows the per-sample MRA distributions; \cref{fig:size2d} shows
why the 2D baseline is unreliable on object size.

\FloatBarrier
\section{Tool-Call Step Analysis}
\label{sec:suppl_steps}

\begin{figure}[t]
  \centering
  \includegraphics[width=\linewidth]{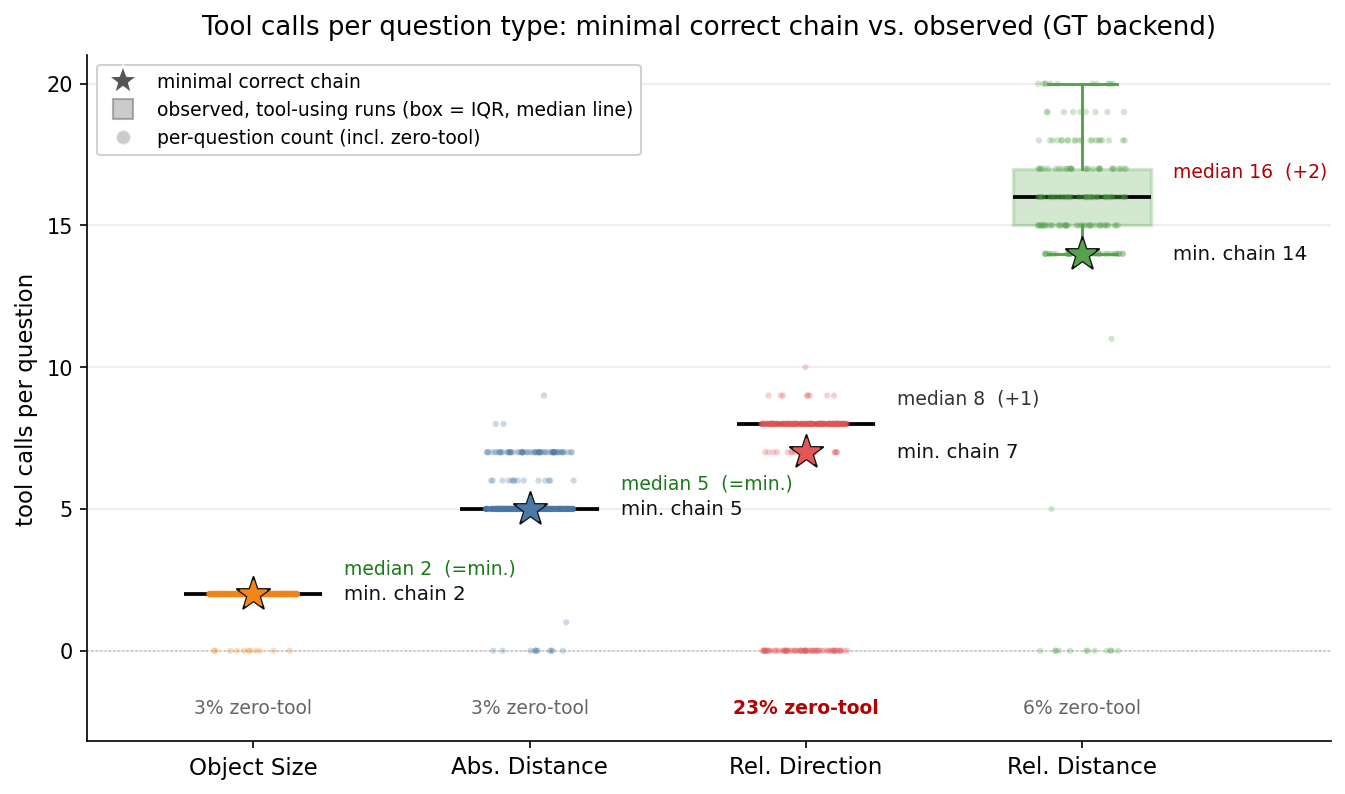}
  \caption{Minimal correct chain vs.\ observed tool-call count per question type (GT backend). Stars mark the minimal chain ($2n_{\text{obj}}+n_{\text{meas}}$, structural per question); boxes summarize the tool-\emph{using} runs (median line, IQR), and jittered points show every run including the zero-tool band (annotated \%). Numeric tasks run the minimal chain exactly; relative direction runs $+1$ (a premature \texttt{relative\_direction} call that errors and is retried) and relative distance $+2$ (redundant distance calls). The $23\%$ zero-tool rate on relative direction is the orchestration \emph{discipline} gap, separate from this overhead.}
  \label{fig:stepanalysis}
\end{figure}

\Cref{fig:stepanalysis} compares the \emph{minimal correct chain} against the
tool calls the orchestrator actually made (GT backend, so the counts isolate
orchestration from perception). The minimal chain is structural,
$2n_{\text{obj}}+n_{\text{meas}}$ for the OBB pipelines (one \texttt{detect} and
one \texttt{project} per object, plus the measurements), dropping the
\texttt{project} step for object size: $2$ calls for size, $5$ for absolute
distance, $7$ for relative direction, and $14$ for relative distance (reference
$+$ $4$ candidates). Three patterns stand out.

\emph{Numeric tasks run the minimal chain exactly.} Object size sits at $2$
($412/423$ samples) and absolute distance at a median of $5$. This near-zero
orchestration overhead is what lets GT-backend accuracy sit within $3$ points of
the tool ceiling (\cref{sec:exp}).

\emph{The multi-object MC tasks carry a small, systematic overhead.} Relative
direction runs $+1$: $198$ of the $223$ engaged runs call
\texttt{relative\_direction} \emph{twice}, once prematurely (before the boxes are
lifted, which returns the ``project first'' error) and again after projecting.
Relative distance runs $+2$, from redundant \texttt{calculate\_object\_distance}
calls beyond the $4$ it needs. Both are benign, since the answer is still correct,
but sequencing is not tight on the longer chains.

\emph{The discipline failure is a separate zero-tool spike.} Every task carries a
band of runs that call \emph{no} tool and answer from the frames: $2.6$--$5.6\%$
on the numeric and relative-distance tasks, but $23\%$ ($67/290$) on relative
direction. Because that unaided guess is no better than chance ($25.9\%$ visual
vs.\ $29.3\%$, \cref{sec:exp}), it is this spike rather than the sequencing
overhead that caps relative-direction accuracy at $80\%$ against a $100\%$
geometric ceiling.

\FloatBarrier
\section{Relative-Direction Subtype Breakdown}
\label{sec:suppl_reldir}

\Cref{tab:reldir_subtype} breaks relative-direction accuracy down by subtype.
The GT oracle is stable across subtypes ($73.6$--$84.1\%$); WildDet3D trails it
by at most $8$~pt everywhere except the ``backward hard'' cases
($84.1\to 68.1\%$), the doubly boundary-sensitive ones: they are 4-way quadrant
questions, so center noise can flip the target across the $90^\circ$ front/back
split, and they use the \texttt{away} facing mode, which negates $\mathbf{f}$, so
noise on the \emph{viewer} center swings the bearing too (\cref{sec:suppl_tools}).
Even so, the overall oracle gap is only $6.6$~pt against a $25$-pt gap on relative
distance. The bearing reads object \emph{centers} alone and never their extents,
which makes it far less detector-sensitive: swapping GT boxes for WildDet3D costs
the tool-only ceiling $11$~pt ($100.0\to 89.0\%$), where the same swap costs
relative distance $25.6$~pt ($94.9\to 69.3\%$). The larger share of the remaining
error on this task is therefore tool-use discipline rather than perception.

% [H] keeps the table inside Sec. G instead of floating it past the section end.
\begin{table}[H]
  \centering\footnotesize
  \setlength{\tabcolsep}{6pt}
  \begin{tabular}{@{}lccc@{}}
    \toprule
    Subtype & N & GT Oracle & WildDet3D \\
    \midrule
    Backward Easy (3-way) & 53 & 79.2\% & 75.5\% \\
    Backward Hard (4-way) & 69 & \textbf{84.1\%} & 68.1\% \\
    Forward Easy (3-way)  & 96 & 82.3\% & 74.0\% \\
    Forward Hard (4-way)  & 72 & 73.6\% & \textbf{76.4\%} \\
    \midrule
    \emph{Overall} & 290 & \textbf{80.0\%} & 73.4\% \\
    \bottomrule
  \end{tabular}
  \caption{Relative-direction accuracy by subtype (``easy'' = 3-way left/right/back; ``hard'' = 4-way quadrant; ``backward'' = the \texttt{away} facing mode). The GT oracle averages $6.1$ tool calls per question over all runs, or $8.0$ (median $8$) over the runs that call any tool at all; the gap to the $100\%$ geometric ceiling is the $23\%$ zero-tool guessing rate.}
  \label{tab:reldir_subtype}
\end{table}

\FloatBarrier
\section{Autonomous Probe: Tool Feedback and Prompting}
\label{sec:suppl_auto}

The no-recipe probe of \cref{sec:exp} strips the per-task recipe and leaves only
the generic tool description of \cref{lst:hint-auto}: the model must route and
order the tools itself. This section reports two runs, both on the GT backend.
The first is a staged ablation on $80$ questions (absolute distance and object
size only) that varies \emph{what the environment tells the model} and produced
the tool-error strings we now ship (\cref{tab:auto_stages}). The second is the
four-task probe on $160$ questions, which supplies the numbers reported in
\cref{tab:autonomous}. They are separate runs over different task mixes and
question samples, so their per-task numbers are not directly comparable; the
four-task probe is the one we report.

\paragraph{Staged ablation (80 questions).}
\emph{Stage 1 (terse errors)} exposes the tool errors of the
original library. Calling \texttt{calculate\_object\_distance} before detecting
and lifting the boxes returns only \emph{``No 3D box found for [tv, sink]''},
which does not say what to do next, so the model retries the same call: $138$
error results over $80$ questions, and a mean of $5.0$ calls where $3.5$ suffice.
This is the looping the main text reports, and it costs the most on distance
($0.58$ MRA), the task with the longest chain.

\emph{Stage 2 (actionable tool errors)} rewrites every tool error into an actionable
instruction naming the missing step, e.g.\ \emph{``No world-space box for [tv,
radiator]. For each of these objects call \texttt{project\_box\_to\_world}
first.''} Errors collapse to $36$ and distance jumps to $0.88$, but the prompt
still never shows the model \emph{when} a tool is needed: zero-tool guessing rises
to $14/80$, dragging size down to $0.74$.

\emph{Stage 3 (few-shot order examples)} appends \cref{lst:hint-fewshot}: one
tool-\emph{order} example per task, showing which tools to chain and in what order, never
an answer. Both failure modes disappear at once (errors $13$, zero-tool $0/80$)
and MRA reaches $0.93$. Within this ablation the examples look decisive; the
four-task probe below shows they are not.

% [H] pins the stage table above the listing it is discussed with, as elsewhere.
\begin{table}[H]
  \centering\footnotesize
  \setlength{\tabcolsep}{4pt}
  \begin{tabular}{@{}lccccc@{}}
    \toprule
    Condition & MRA & Dist. & Size & Errors & Zero-tool \\
    \midrule
    1. Terse tool errors      & 0.729 & 0.575 & 0.883 & 138 & 5/80 \\
    2. \, + actionable errors & 0.800 & 0.878 & 0.723 & 36  & 14/80 \\
    3. \, + few-shot examples & \textbf{0.933} & \textbf{0.938} & \textbf{0.928} & \textbf{13} & \textbf{0/80} \\
    \bottomrule
  \end{tabular}
  \caption{The staged ablation on tool feedback (80 questions: 40 absolute distance, 40 object size; GT backend). \emph{Errors} counts tool results returned as errors across all $80$ traces; \emph{Zero-tool} counts runs that answered without calling a single tool. Better feedback removes the looping, and here the few-shot examples remove the remaining guessing; on the four-task probe (\cref{tab:autonomous}) they no longer help.}
  \label{tab:auto_stages}
\end{table}

\begin{lstlisting}[style=prompt,caption={Few-shot block appended to the \texttt{autonomous\_3d} hint (\cref{lst:hint-auto}) in stage 3. One tool-order example per task; no ground-truth answers and no per-question recipe.},label={lst:hint-fewshot}]
Examples of tool choice+order (these show which
tools, NOT the answers):
- size: 'longest dimension of the chair?'
    -> detect_object_3d(chair);
       calculate_object_size(chair).
- distance: 'distance between tv and sofa?'
    -> detect_object_3d(tv);
       project_box_to_world(tv);
       detect_object_3d(sofa);
       project_box_to_world(sofa);
       calculate_object_distance(tv, sofa).
- rel-distance: 'which of X/Y/Z is closest to the
  lamp?' -> lift the lamp and each candidate
       (detect_object_3d + project_box_to_world),
       calculate_object_distance(lamp, each),
       pick the smallest.
- rel-direction: 'standing by A facing B, where
  is C?' -> lift A, B, C (detect_object_3d +
       project_box_to_world each),
       relative_direction(A, B, C).
\end{lstlisting}

\paragraph{Four-task probe (160 questions).}
Running all four tasks at once ($40$ questions each), with the rewritten tool
errors in place, gives the results of \cref{tab:autonomous} and confirms that
routing itself is not the bottleneck. With the \emph{generic} prompt alone (no
recipe, no examples), the model calls the correct final tool on $98\%$ (size),
$100\%$ (distance) and $90\%$ (relative distance) of questions and scores $0.97$
MRA, $0.94$ MRA and $92.5\%$ accuracy, matching its per-task numbers under the
hardcoded recipe. Relative direction is the single exception ($70\%$ routed,
$70.0\%$ accuracy): on $30\%$ of those questions it calls no tool at all, the
same discipline failure that caps the recipe-driven runs (\cref{sec:suppl_steps}).

The few-shot examples that were decisive in the staged ablation do \emph{not}
carry over. Repeating the four-task probe with the stage-3 few-shot block leaves
size ($1.00$) and relative direction ($75.0\%$) slightly better but drags
absolute distance to $0.89$ and relative distance to $80.0\%$, below the generic
prompt on both. Once the tools report their errors clearly, the in-context
examples add nothing to routing and can bias the model toward the example's chain
on questions that need a different one. We therefore report the generic prompt as
the headline autonomous condition, not the few-shot one.